\documentclass[conference]{IEEEtran}
\IEEEoverridecommandlockouts

\usepackage{graphicx}
\usepackage{multirow}
\usepackage{subcaption}
\usepackage{amsmath}
\usepackage{nccmath}
\usepackage{makecell}
\usepackage{float}
\usepackage{verbatim}

\usepackage{cite}
\usepackage{amsmath,amssymb,amsfonts}
\usepackage{algorithmic}
\usepackage{graphicx}
\usepackage{textcomp}
\usepackage{xcolor}
\usepackage{multirow}
\def\BibTeX{{\rm B\kern-.05em{\sc i\kern-.025em b}\kern-.08em
    T\kern-.1667em\lower.7ex\hbox{E}\kern-.125emX}}

\usepackage{float}
\floatstyle{plaintop}
\restylefloat{table}
\usepackage{array}
\newcommand{\PreserveBackslash}[1]{\let\temp=\\#1\let\\=\temp}
\newcolumntype{C}[1]{>{\PreserveBackslash\centering}p{#1}}
\newcolumntype{R}[1]{>{\PreserveBackslash\raggedleft}p{#1}}
\newcolumntype{L}[1]{>{\PreserveBackslash\raggedright}p{#1}}

\def\BibTeX{{\rm B\kern-.05em{\sc i\kern-.025em b}\kern-.08em
    T\kern-.1667em\lower.7ex\hbox{E}\kern-.125emX}}

\usepackage{booktabs}

\usepackage{hyperref}
\hypersetup{
    colorlinks=true,
    allcolors=blue
}

\definecolor{ckgreen}{rgb}{0,0.56,0}

\makeatletter
\let\old@ps@IEEEtitlepagestyle\ps@IEEEtitlepagestyle
\def\confheader#1{%
    \def\ps@IEEEtitlepagestyle{%
        \old@ps@IEEEtitlepagestyle%
        \def\@oddhead{\strut\hfill#1\hfill\strut}%
        \def\@evenhead{\strut\hfill#1\hfill\strut}%
    }%
    \ps@headings%
}
\makeatother

\IEEEpubid{
\begin{minipage}[t]{\textwidth}\ \\[10pt]
      \small{
      (Copyright $\copyright$ 2022 IEEE. This work has been submitted to the IEEE for possible publication. Copyright may be transferred without notice, after which this version may no longer be accessible.)}
\end{minipage}
}

\begin{document}

\title{Performance Analysis of YOLO-based Architectures for Vehicle Detection from Traffic Images in Bangladesh}

%%%%%%% Author Info %%%%%%%%%%%%%
\author{ 
    \IEEEauthorblockN{Refaat Mohammad Alamgir, Ali Abir Shuvro, Mueeze Al Mushabbir, Mohammed Ashfaq Raiyan, Nusrat Jahan Rani, \\
    Md. Mushfiqur Rahman, Md. Hasanul Kabir, and Sabbir Ahmed\\}

    \IEEEauthorblockA{Department of Computer Science and Engineering,\\Islamic University of Technology, Gazipur, Bangladesh\\}
    
    \IEEEauthorblockA{Email: \{refaatalamgir, aliabir, almushabbir, ashfaqraiyan, nusratjahan44, \\
    mushfiqur11, hasanul, sabbirahmed\}@iut-dhaka.edu}
}

\maketitle

\begin{abstract}

The task of locating and classifying different types of vehicles has become a vital element in numerous applications of automation and intelligent systems ranging from traffic surveillance to vehicle identification and many more. In recent times, Deep Learning models have been dominating the field of vehicle detection. Yet, Bangladeshi vehicle detection has remained a relatively unexplored area. One of the main goals of vehicle detection is its real-time applicat ion, where `You Only Look Once' (YOLO) models have proven to be the most effective architecture. In this work, intending to find the best-suited YOLO architecture for fast and accurate vehicle detection from traffic images in Bangladesh, we have conducted a performance analysis of different variants of the YOLO-based architectures such as YOLOV3, YOLOV5s, and YOLOV5x. The models were trained on a dataset containing 7390 images belonging to 21 types of vehicles comprising samples from the DhakaAI dataset, the Poribohon-BD dataset, and our self-collected images. After thorough quantitative and qualitative analysis, we found the YOLOV5x variant to be the best-suited model, performing better than YOLOv3 and YOLOv5s models respectively by 7 \& 4 percent in mAP, and 12 \& 8.5 percent in terms of Accuracy.
\end{abstract}


\section{Introduction}
The process of locating vehicles and identifying their types is known as vehicle detection. In this era of automation, where the world is quickly moving towards autonomous vehicles, developing fast and accurate vehicle detection algorithms has become a demand of time. Vehicle detection is vital for traffic surveillance, %traffic engineering,
vehicle counting, vehicle speed measurement, identification of traffic accidents, traffic flow prediction, and numerous other applications\cite{hadi2014vehicle, mandal2020object, chintalacheruvu2012video, Ashrafee_2022_WACV}. %chen2014vehicle
%The Intelligent Transportation Systems (ITS) use vehicle detection to improve public safety, reduce congestion, upgrade travel and transit information, and reduce detrimental environmental impacts \cite{chintalacheruvu2012video}.
% generate cost savings, 

Vehicle detection is a subdomain of the object detection domain. In both cases, the main goal is identifying target objects from images or videos, and then correctly classifying them among the defined classes. In vehicle detection tasks, these defined classes are strictly restricted to vehicle types. 
%Therefore, object detection algorithms can directly be used for vehicle detection tasks like vehicle counting \cite{mandal2020object}, vehicle detection from satellite images \cite{chen2014vehicle}, etc. 
Compared to traditional algorithms for Computer Vision, Deep Learning (DL) models tend to perform much better due to their ability of finding patterns and relationships among patterns that seem to be hidden from human perception \cite{ashikur2022twoDecades}. 
Researchers have obtained great results by using popular object detection algorithms like Faster R-CNN \cite{fan2016closer}, %ren2015faster
YOLO \cite{zhou2016image}, %Redmon_2016_CVPR, corovic2018real
% YOLO9000 \cite{Redmon_2017_CVPR}, 
YOLOv3 \cite{zhang2019vehicle} %redmon2018yolov3
, YOLOv5 \cite{kasper2021detecting,rahman2021densely} etc. in vehicle detection. 
%An automated traffic system is an essential requirement satisfied by vehicle detection methods. 
% IMPORTANT>>
%In this particular problem, speed is a major factor where real-time image detection and classification are necessary. Keeping these in mind, we have performed our experiment on YOLO models which are best known for their faster performances and one-step approach. 

All of the difficulties presented by a conventional object detection problem exist for vehicle detection tasks. Dual priority (between object localization and classification), speed (for real-time detection), class imbalance, and limited data are some of these significant challenges in object detection \cite{fessel_2019}. Besides, vehicle detection tasks face a few additional obstacles as well. Firstly, the physical distinction between different classes of vehicles is minimal, making the classification task extremely difficult. Secondly, the high number of vehicles per image and their partial occlusions make accurate localization quite strenuous. 
Recent developments in Deep Learning architectures can tackle many of these challenges very well. Since most use cases of vehicle detection require real-time detection, complex and expensive algorithms become ill-suited for such tasks. 

So, keeping these in mind, we have performed our experiment on models based on You-Only-Look-Once (YOLO) architectures, which are best known for their faster performances and one-step approach.
Hence, the goal of this research is to find a balanced YOLO-based vehicle detection algorithm that has a good trade-off between accuracy and speed.
We have worked to illustrate the impacts of different YOLO-based state-of-the-art models of object detection at present in the field of vehicle detection. We determined which model would give us the highest accuracy and also which model would be perfect considering a real-time implementation. Our contributions to this paper are as follows:
\begin{itemize}
    \item 
 We provided a thorough performance evaluation of different YOLO-based vehicle detection algorithms, such as: YOLOv3, YOLOv5s, and YOLOv5x. 
    \item
 We curated a dataset majorly from the DhakaAI dataset \cite{DhakaAI2020} that was further enriched with another dataset PoribohonBD \cite{tabassum2020poribohon} and some data of our own. For better performance, we merged that dataset with some other data collected from multiple sources. 
    \item
    Finally, we provided an extensive qualitative and quantitative analysis to compare the performance of each of the models.
\end{itemize}

The rest of this paper is organized as follows. In Section \ref{sec:literatureReview}, we briefly review the existing object detection algorithms used for vehicle detection. Section \ref{sec:methodology} is comprised of discussions on the datasets and metrics for this task. Extensive experiments are performed and comparative analysis is shown in Section \ref{sec:results} to evaluate the state-of-the-art methods. Finally, we conclude this paper and discuss further work in Section \ref{sec:conclusion}.

\section{Literature Review}\label{sec:literatureReview}

Object detection is a crucial task in computer vision systems. With its wide range of applications, including vehicle detection, object detection is one of the most studied topics among computer vision researchers. With the advent of deep learning algorithms along with the availability of high-end computational power, modern object detection algorithms are much faster and superior to their predecessors. Before 2014, almost all algorithms used traditional computer vision theories to tackle object detection tasks. 
Scale Invariant Feature Transform (SIFT) \cite{lowe1999object} algorithm was one of the first algorithms that could give reasonable accuracy while maintaining high speed. Later on, the Viola-Jones Detectors (VJ Det.) algorithm was developed with a sliding window approach \cite{viola2001rapid}. %viola2004robust.
The Histogram Oriented Gradients Detectors (HOG Det.) \cite{dalal2005histograms} and the Deformable Part-based Model (DPM) \cite{felzenszwalb2009object} 
%felzenszwalb2010cascade, felzenszwalb2008discriminatively
were the most noteworthy ones among them. Even some of the modern papers embraced these approaches \cite{wei2019multi,neumann2017online}. The traditional approaches are still relevant in many cases and their effective use can improve the performance of modern algorithms. Many advanced models borrow insights like hard negative mining, bounding box regression, etc., from these traditional methods.

In recent years, deep learning has paved the way for fast and accurate classification and object detection algorithms~\cite{ahmed2022lessIsMore}. These algorithms can be generally classified into two broad categories - the two-step algorithms and the one-step algorithms. The two-step algorithms \cite{ren2016faster} %ren2015faster
perform the object localization and object classification in two disconnected steps, and the accuracy depends on the models used in each step. The one-step algorithms, on the other hand, combine localization and classification into one task to enhance speed while maintaining decent accuracy. The You Only Look Once (YOLO) algorithm and all its subsequent versions are considered as one-step algorithms. Both types of algorithms are heavily used in vehicle detection. However, the latter one is preferred due to its high speed.

%\subsection{YOLO family}

The You Only Look Once (YOLO) family of algorithms are considered true one-step algorithms. These are unified and real-time object detection algorithms that can detect objects accurately with a single neural network. Since its first release in 2015, it went through several upgrades and several versions of this algorithm were released \cite{thuan2021evolution}. Considering its speed while maintaining high enough accuracy, many consider it to be the best object detection algorithm. 

The YOLOv1 \cite{Redmon_2016_CVPR} based on the Darknet architecture was its first major improvement on the core YOLO architecture. Afterwards, the YOLOv2 to YOLOv5 were subsequently developed \cite{Diwan2022}.
% \cite{Redmon_2017_CVPR,redmon2018yolov3,bochkovskiy2020yolov4,glenn_jocher_2021_4679653}. 
Among these architectures, the latest one, i.e., the YOLOv5 model is currently the best-performing model. The architectures of YOLOv4 and YOLOv5 are quite similar. The notable difference in YOLOv5 is its shape.

\section{Methodology} \label{sec:methodology}

\subsection{Dataset}
The performance of a deep learning algorithm heavily relies on the quality of the dataset. Although many large datasets are readily available for object detection tasks; the unavailability of good datasets is the biggest bottleneck for vehicle classification. And even when some vehicle detection datasets are available they suffer from lacking generalization for diversified street types and class imbalance.

\begin{figure}[t]
    \centering
    \vspace*{-1.2cm}
    \includegraphics[width=1\columnwidth]{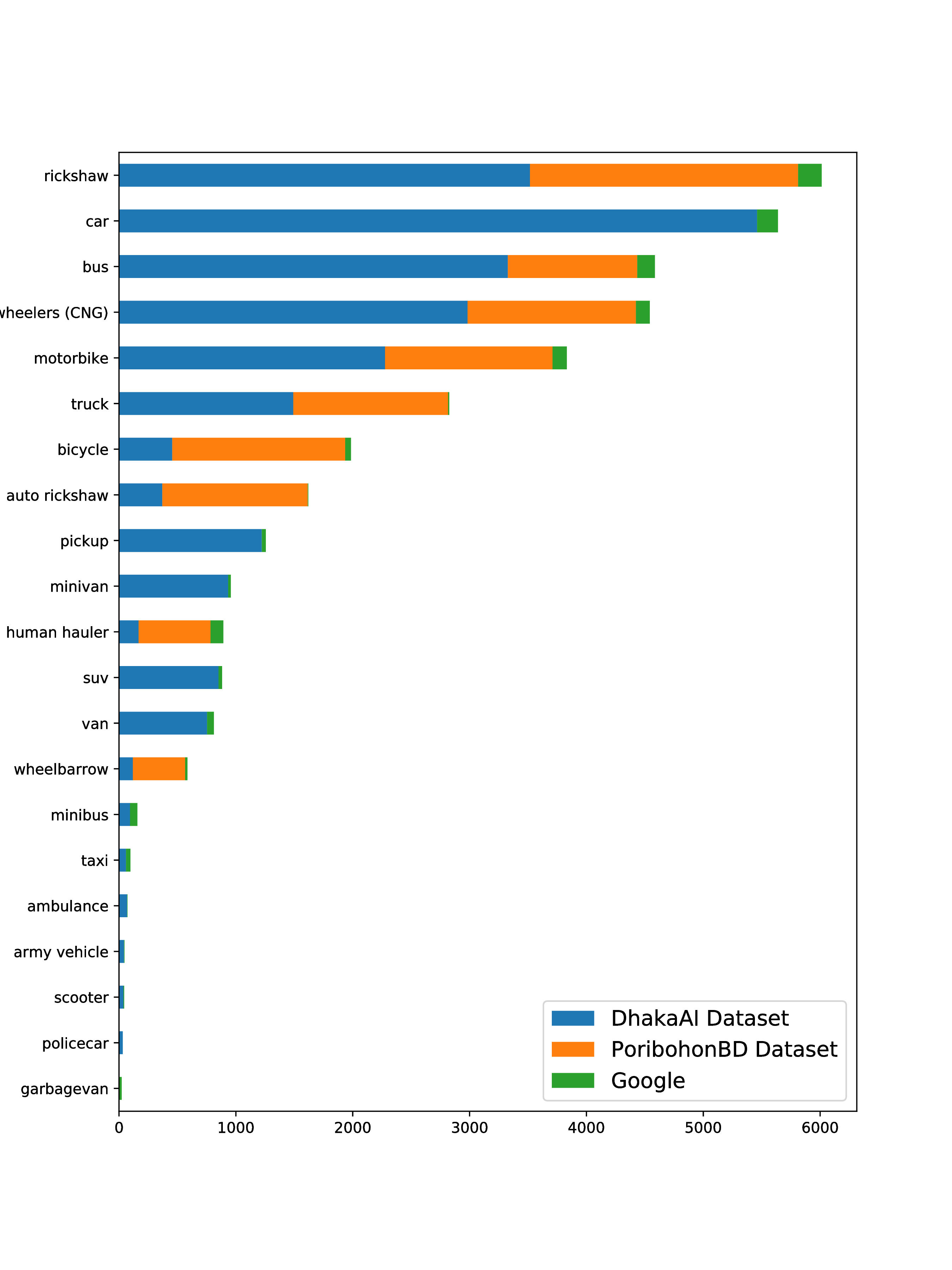}
    \caption{Distribution of images of the combined dataset}
    \label{fig:dataset}
    
\end{figure}

To perform vehicle detection for Bangladeshi vehicles, we used DhakaAI final training dataset~\cite{DhakaAI2020}. The dataset is largely skewed with some of the classes appearing much less than others. This affected the performance of the model so other sources of data were necessary. We collected the Poribohon-BD dataset \cite{tabassum2020poribohon} and the samples from the Internet to offer balance to the dataset. We divided the DhakaAI dataset into 2767 and 223 images for training and validation sets respectively, and similarly divided the Poribohon-BD dataset into 4486 and 355 images, and the collected images dataset into 137 and 11 images respectively. The DhakaAI Final Training Dataset for Round 1 was used for the Test set which consists of 499 images. As shown in Figure~\ref{fig:dataset} the dataset contains images of 21 different types of vehicles: ambulance, auto-rickshaw, bicycle, bus, car, garbage van, human hauler, minibus, minivan, motorbike, pickup, army vehicle, police car, rickshaw, scooter, SUV, taxi, three-wheelers (CNG), truck, van, wheelbarrow. The combined dataset contains images in different camera angles and environment variations like night images, and low light images captured using different cameras.

\subsection{Model Description}
The model that we are primarily interested in for our requirement is YOLOv5 which has three parts; the Backbone, Neck, and Head. The Backbone is formed of the CSP Network and the Focus Structure. The Neck holds the PANet and SPP Block while the Head is similar to other YOLO models like YOLOv3 in addition to a GIoU-loss. This model outperforms all the YOLO versions in terms of mAP and speed~\cite{liu2021performance}.
\begin{figure*}[htbp]
    \centering
    \includegraphics[width=0.99\linewidth]{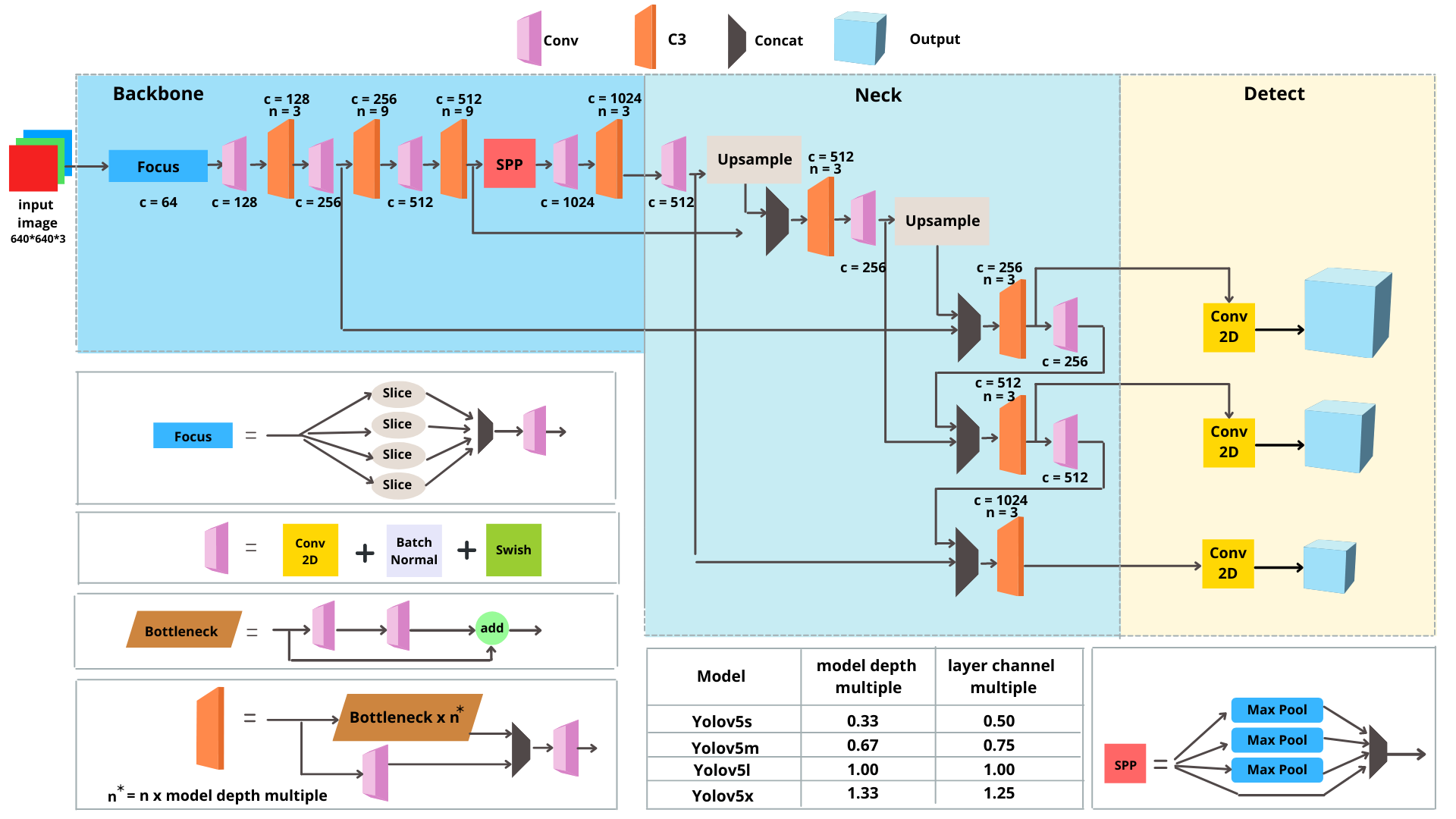}
    \caption{Network architecture of Yolov5} 
    % It consists of three parts: (1) Backbone, (2) Neck, and (3) Detection Layer.}
    \label{fig:yolov5}
\end{figure*}

The YOLOv5 network consists of three pieces, namely, Backbone, Neck, and Head. The Backbone part consists of the Focus module, convolution module, C3 module, and SPP module. The Focus module contains the output of each Channel and then uses a standard convolution module to output. The C3 module contains three standard convolution layers and multiple Bottleneck modules. The number of Bottlenecks is determined by the model depth multiple. 
The Neck part generates a feature pyramid. Finally, from the feature pyramid, Head Layer outputs detection results (class probabilities, confidence scores, and bounding boxes). To train YOLOv5, four versions have been proposed- YOLOv5s (small), YOLOv5m (medium), YOLOv5l (large), and YOLOv5x (extra-large). Among these 4 versions of YOLOv5, YOLOv5s is the fastest while YOLOv5x gives the fastest and most accurate performance with the highest mAP value. Figure~\ref{fig:yolov5} gives a detailed overview of the Yolov5 architecture. 
% \cite{zhou2021safety}.

{Sections/4-result}
\section{Results and Discussion}    \label{sec:results}
%result and discussion, 4a- experimental setup, 4b- evaluation metrics, 4c- quantitative analysis, 4d- Qualitative analysis 4e- Error analysis (optional)
%merge both tables

\subsection{Experimental Setup}

The experiment was run in a Python environment using Google Colaboratory. 
% Our combined dataset was uploaded and was used to train all the different models (YOLOv3, YOLOv5s, YOLOv5x).
The average usable memory of the machine was 12GB and the average disk space used was around 108GB. 
The sample images were randomly split into 60\% for training, 20\% for validation, and 20\% for testing. The batch size was taken as 16 for all the models. This was done because smaller batches being noisy can reduce generalization errors \cite{ahmed2022lessIsMore}. 
% The weights were saved in a Google Drive folder for resumption and future use. 
Model Checkpoints were used to save the model with the best validation accuracy and for a continuation of progress. The convergence of the models had huge dependencies on many of the parameters and the size of the weight parameters. Using the Early Stopping technique, the total time that was required to complete the training of all these models was 563 hours.  

\subsection{Evaluation Metrics}
%For any computer vision project, evaluation is necessary to understand how well the model is performing in the given task. Traffic detection could be classified as an object detection task and consequently, it makes sense to use the appropriate metrics for object detection. And for object detection, the most appropriate one is mAP, which we have used. 

We used Mean Average Precision (mAP) as the main evaluation metric. The predicted bounding box is labeled as true positive if the Intersection-Over-Union (IOU) threshold is satisfied. IOU is a measurement of how much area of the ground truth bounding box has in common with the predicted bounding box, and as the name suggests it is measured as the intersection of the ground truth and the predicted bounding box divided by the union of those two boxes. The predicted bounding box also contains a Confidence score. 

\begin{ceqn}
\begin{align}
    IoU = \frac{GBox \cap PBox}{GBox \cup PBox}
\end{align}
\end{ceqn}

Here, GBox = Ground Truth bounding box, and PBox = Predicted bounding box

According to the IoU and Confidence score, we then find out the True Positives (TP), False Positives (FP), and False Negatives (FN) for each class. Afterward, we calculate Precision and Recall according to the following formulas: 

% \begin{ceqn}
% \begin{align}
%     Precision = \frac{\text{TP}}{\text{TP + FP}}
%     Recall = \frac{\text{TP}}{\text{TP + FN}}
% \end{align}
% \end{ceqn}

\begin{align*} 
Precision &= \frac{\text{TP}}{\text{TP + FP}}\\
    Recall &= \frac{\text{TP}}{\text{TP + FN}}
\end{align*}

% \begin{ceqn}
% \begin{align}
    
% \end{align}
% \end{ceqn}

From the Precision and Recall, we find interpolated precisions, \(p_{interpolated}(r)\), at each level of recall, \(r\), by taking the maximum precision value for that level. 

\begin{ceqn}
\begin{align}
    p_{interpolated}(r) = \max_{\tilde{r}\geq r} p(\tilde{r})
\end{align}
\end{ceqn}

The precision-recall pairs are divided into 11 parts at an interval of 0.1, such as: {0.0, 0.1, \dots, 0.9, 1.0}. The area under the curve of precision vs recall known as AP (Average Precision) can be calculated by this formula as well:

\begin{ceqn}
\begin{align}
    AP = \frac{1}{11} \sum_{r \in \{ 0.0, 0.1, \ldots , 0.9, 1.0 \} } p_{interpolated}(r) 
\end{align}
\end{ceqn}

This AP is for one single class. So, now if we take the arithmetic mean of all of the APs of all the classes, then we finally get our mean Average Precision or mAP.

\subsection{Quantitative Analysis}
As we have primarily focused on the faster model %during the selection phase, 
we have tracked the accuracy of the models during the experimental phase. Each of these models was able to detect the vehicle quite impressively but the rate of success differed.  

% \usepackage{multirow}
\begin{table}[H]
\caption{Quantitative analysis of the YOLO models}
\label{quantitativePerformance}
\begin{tabular}{L{1.2cm} C{1.1cm} C{0.45cm} C{0.7cm} C{0.7cm} C{0.7cm}}
\toprule

\multirow{2}{*}{} & \multirow{2}{*}{\textbf{Params}} & \multicolumn{4}{c}{\textbf{Performance}}                                                                                              \\ \cline{3-6} 
                  &                                  & \multicolumn{1}{C{0.5cm}}{\textbf{mAP}} & \multicolumn{1}{c}{\textbf{Accuracy}} & \multicolumn{1}{c}{\textbf{Precision}} & \textbf{Recall} \\ \midrule
\textbf{YOLOv3}   & 61.9M                            & \multicolumn{1}{C{0.45cm}}{0.212}        & \multicolumn{1}{c}{0.192}             & \multicolumn{1}{c}{0.194}              & 0.442           \\ \midrule
\textbf{YOLOv5s}  & 7.3M                             & \multicolumn{1}{C{0.45cm}}{0.247}        & \multicolumn{1}{c}{0.228}             & \multicolumn{1}{c}{0.242}              & 0.290           \\ \midrule
\textbf{YOLOv5x}  & 87.7M                            & \multicolumn{1}{C{0.45cm}}{0.287}        & \multicolumn{1}{c}{0.313}             & \multicolumn{1}{c}{0.247}              & 0.294           \\ \bottomrule
\end{tabular}

\end{table}

As mentioned in Table~\ref{quantitativePerformance}, the detection capabilities of all the models are quite similar as expected after training. If closely observed, variants of YOLOv5 (YOLOv5s and YOLOv5x) seem to be doing better than the others.

\begin{figure}[tb]
    \centering
    \includegraphics[width=0.9\linewidth]{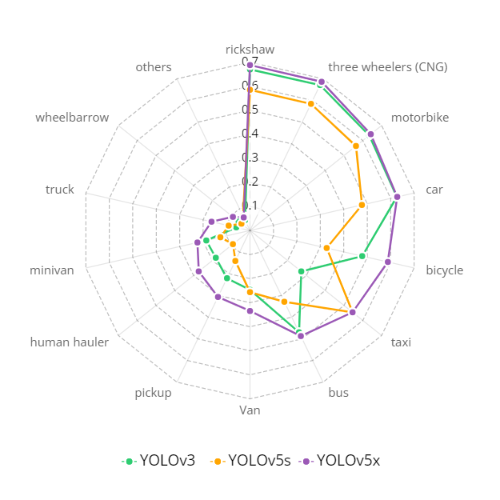}
    \caption{Radar chart comparing the vehicle-wise Average Precision of different models}
    \label{fig:radar}
    \vspace*{-0.5cm}
\end{figure}

If we take a closer look at the radar chart in Figure \ref{fig:radar}, we can see that YOLOv5x accomplished the best average precision across all the classes with mAP of 0.287. It has been seen to perform exceptionally well in the cases of vehicle classes with fewer images. YOLOv5x still outperformed all the other models if the amount of data is larger (seen in the classes like rickshaw, car, motorbike, etc.).

\subsection{Qualitative Analysis}
\begin{figure*}[htbp]
    \centering
    
    % \subfloat[][ \label{...}]
    {\includegraphics[width=0.24\linewidth]{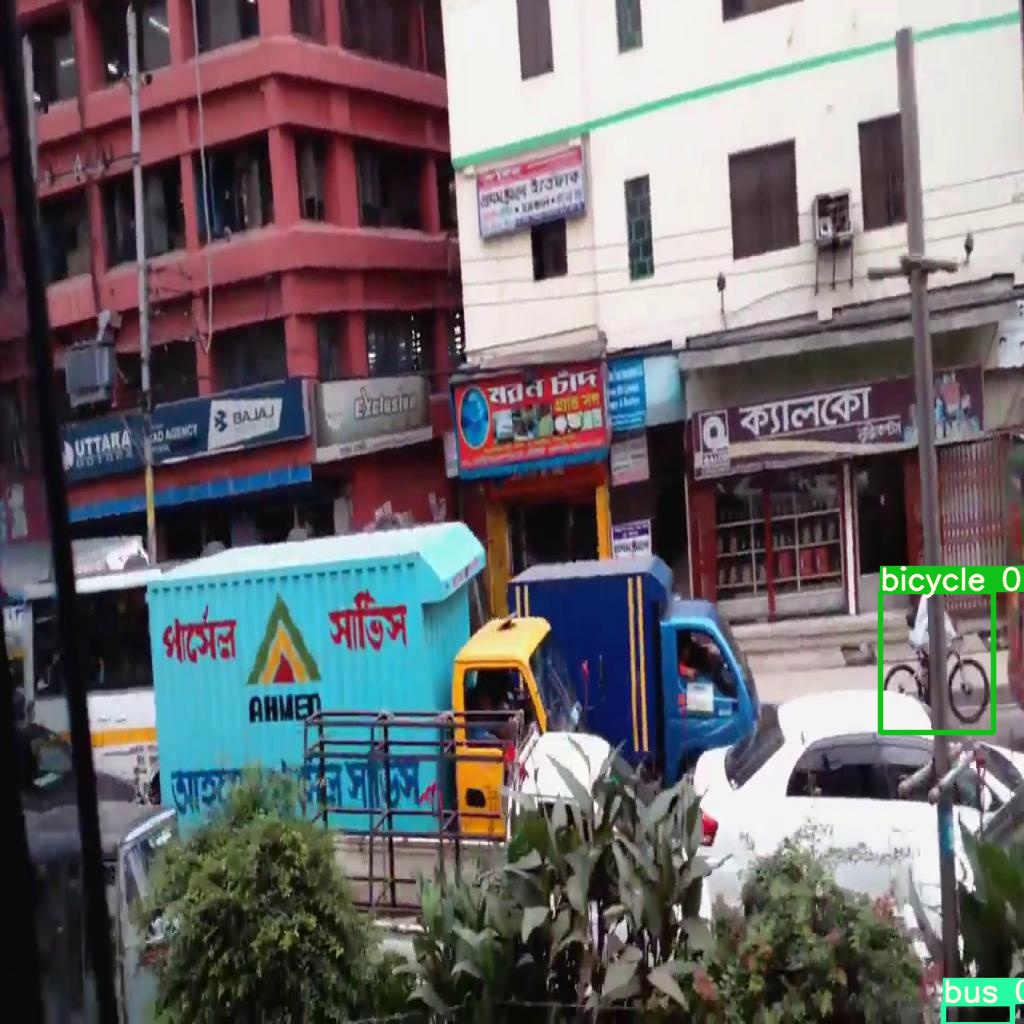}}
    % \subfloat[][ \label{...}]
    {\includegraphics[width=0.24\linewidth]{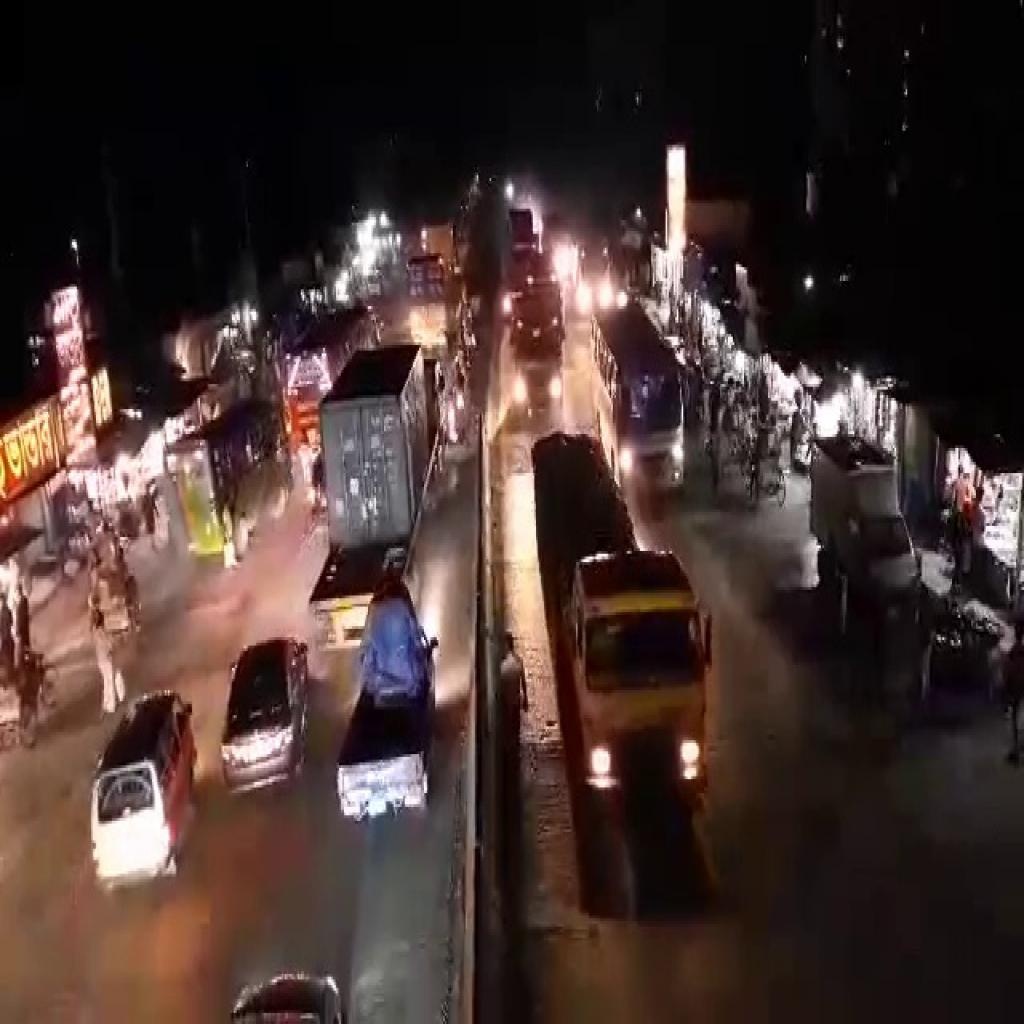}}
    % \subfloat[][ \label{...}]
    {\includegraphics[width=0.24\linewidth]{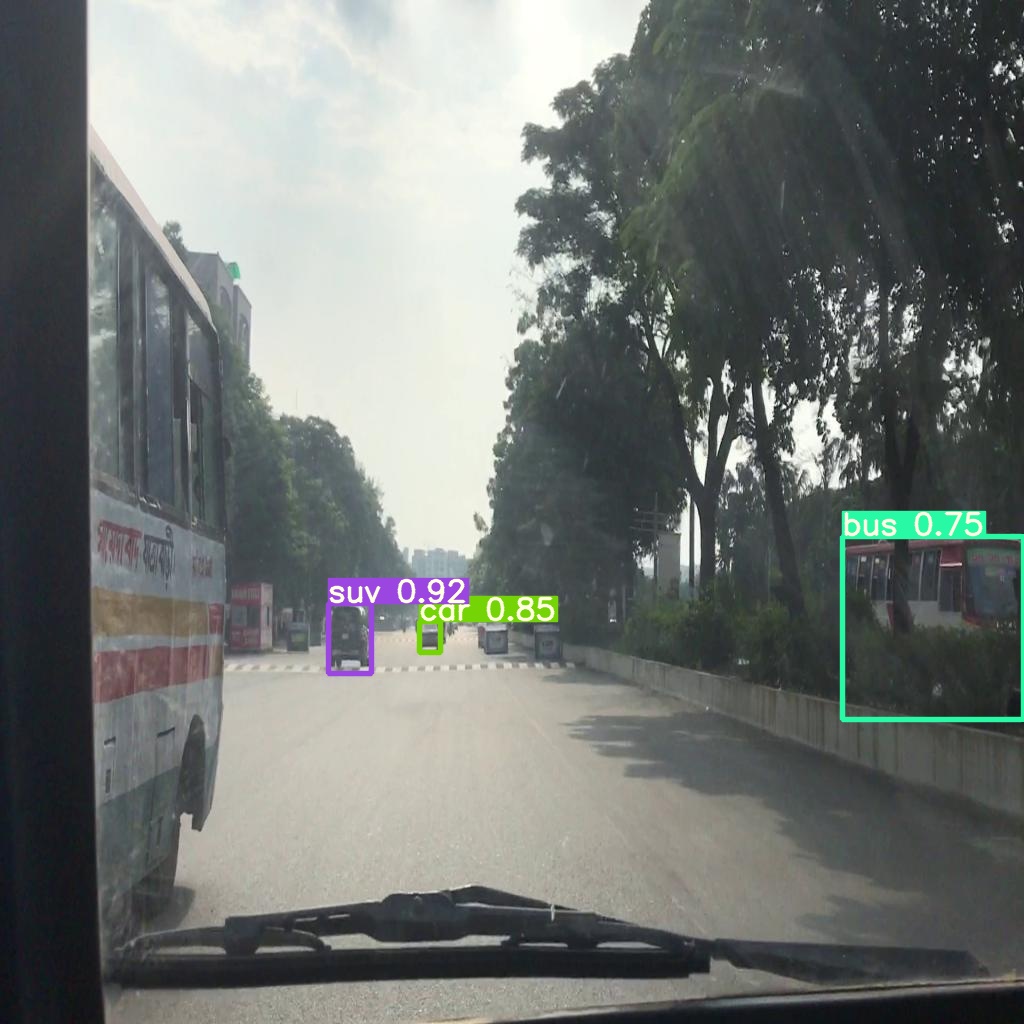}}
    %\subfloat[][YOLOV3 \label{...}]
    {\includegraphics[width=0.24\linewidth]{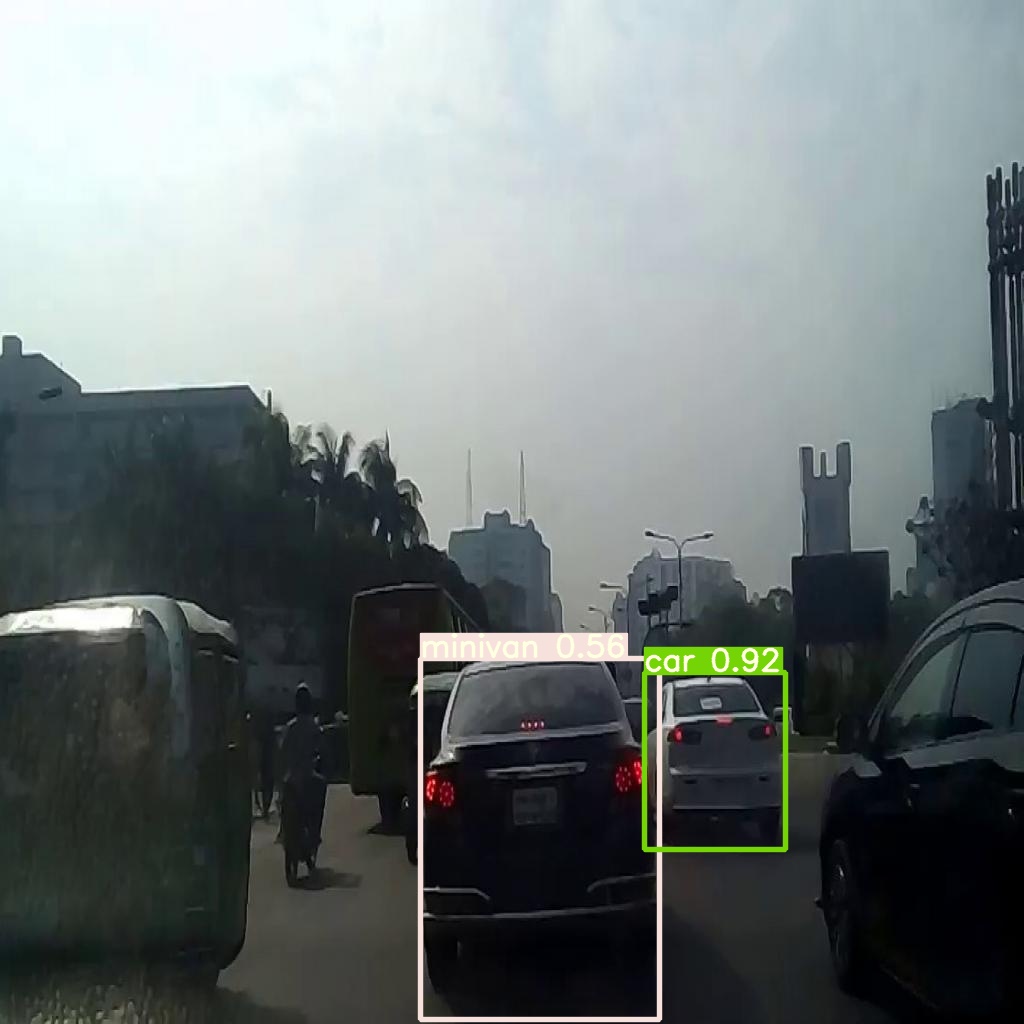}}\\
    
    (a)YOLOv3\\

    \hspace{0.5pt}
    
    % \subfloat[][ \label{...}]
    {\includegraphics[width=0.24\linewidth]{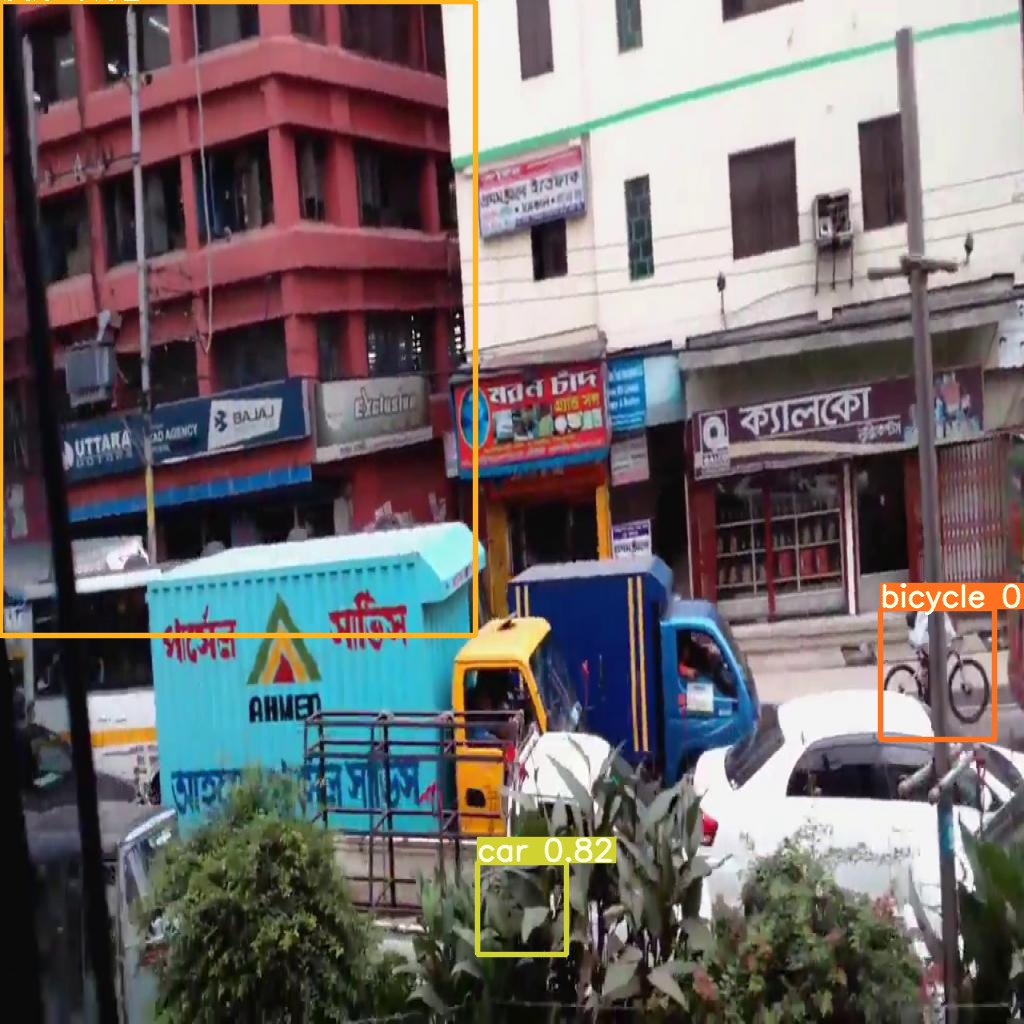}}
    % \subfloat[][ \label{...}]
    {\includegraphics[width=0.24\linewidth]{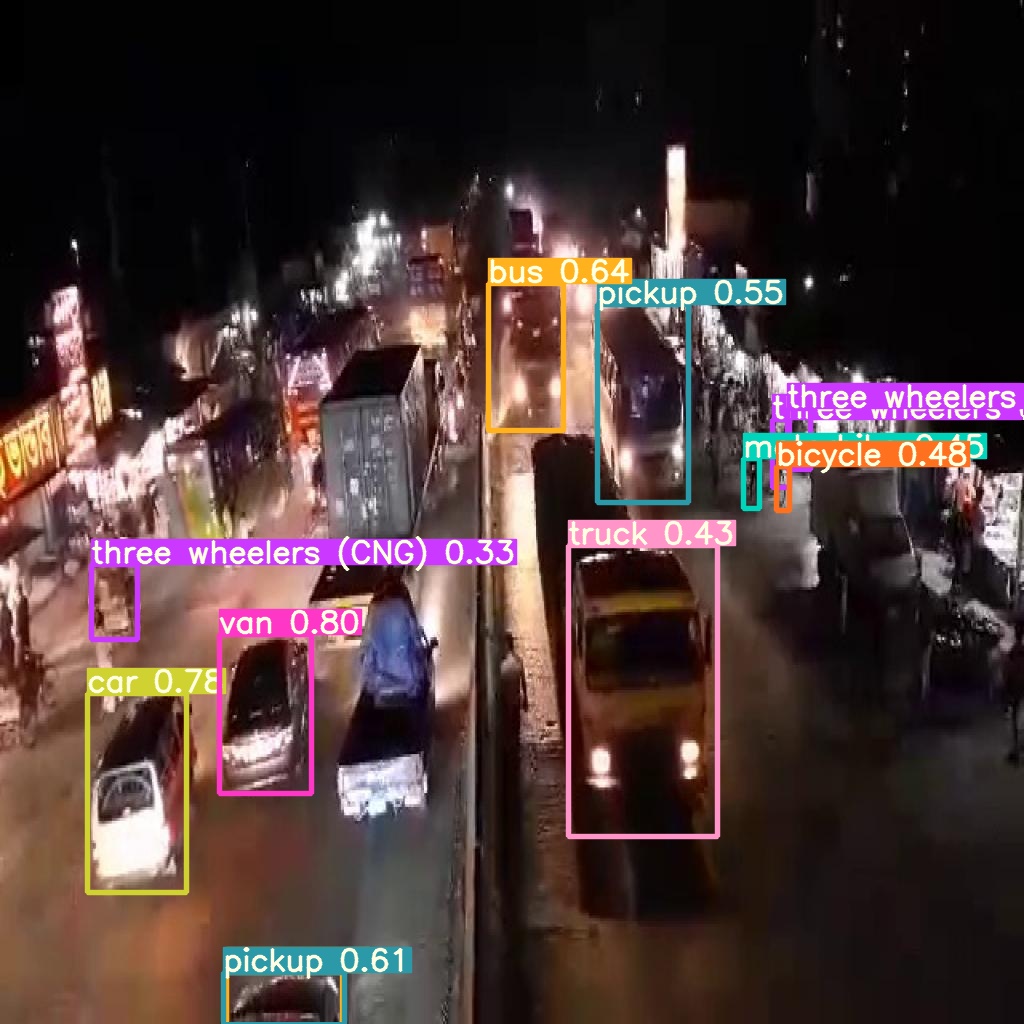}}
    % \subfloat[][ \label{...}]
    {\includegraphics[width=0.24\linewidth]{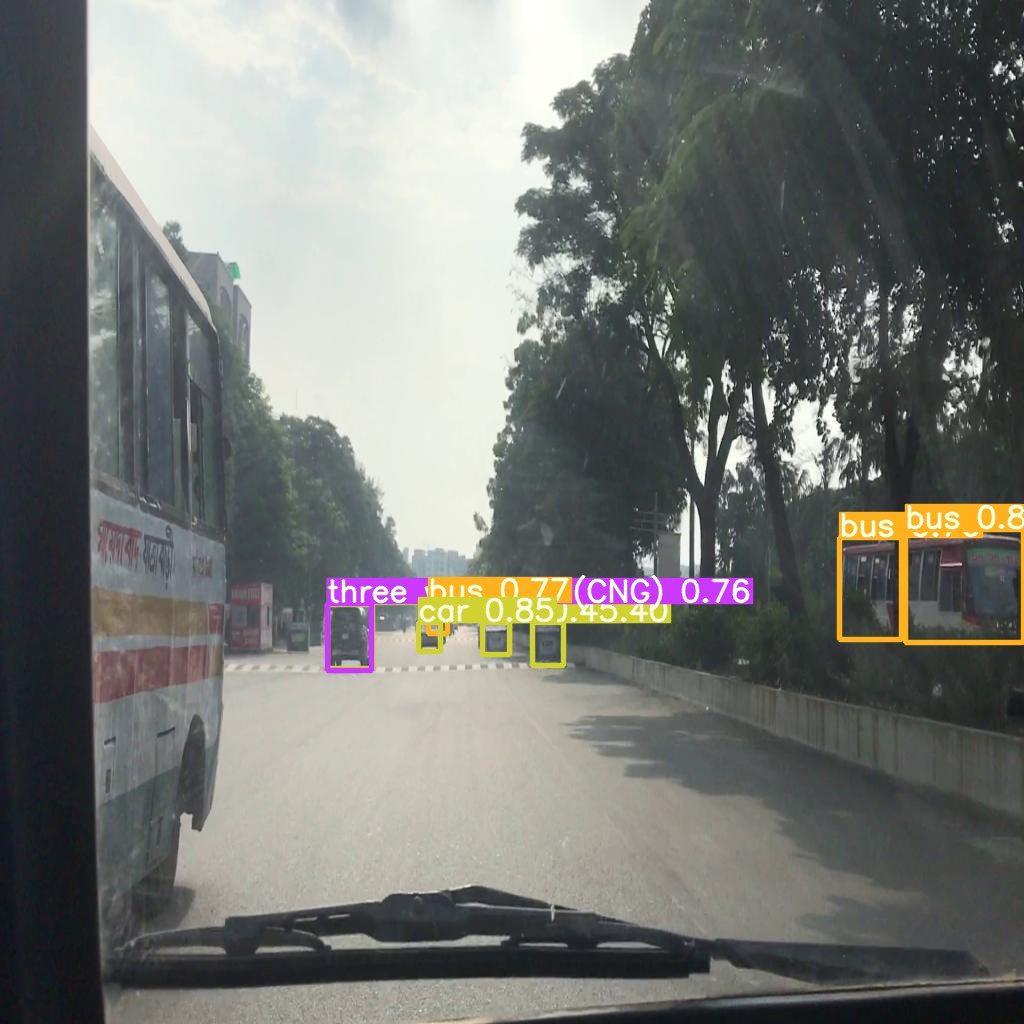}}
    %\subfloat[][YOLOV5S \label{...}]
    {\includegraphics[width=0.24\linewidth]{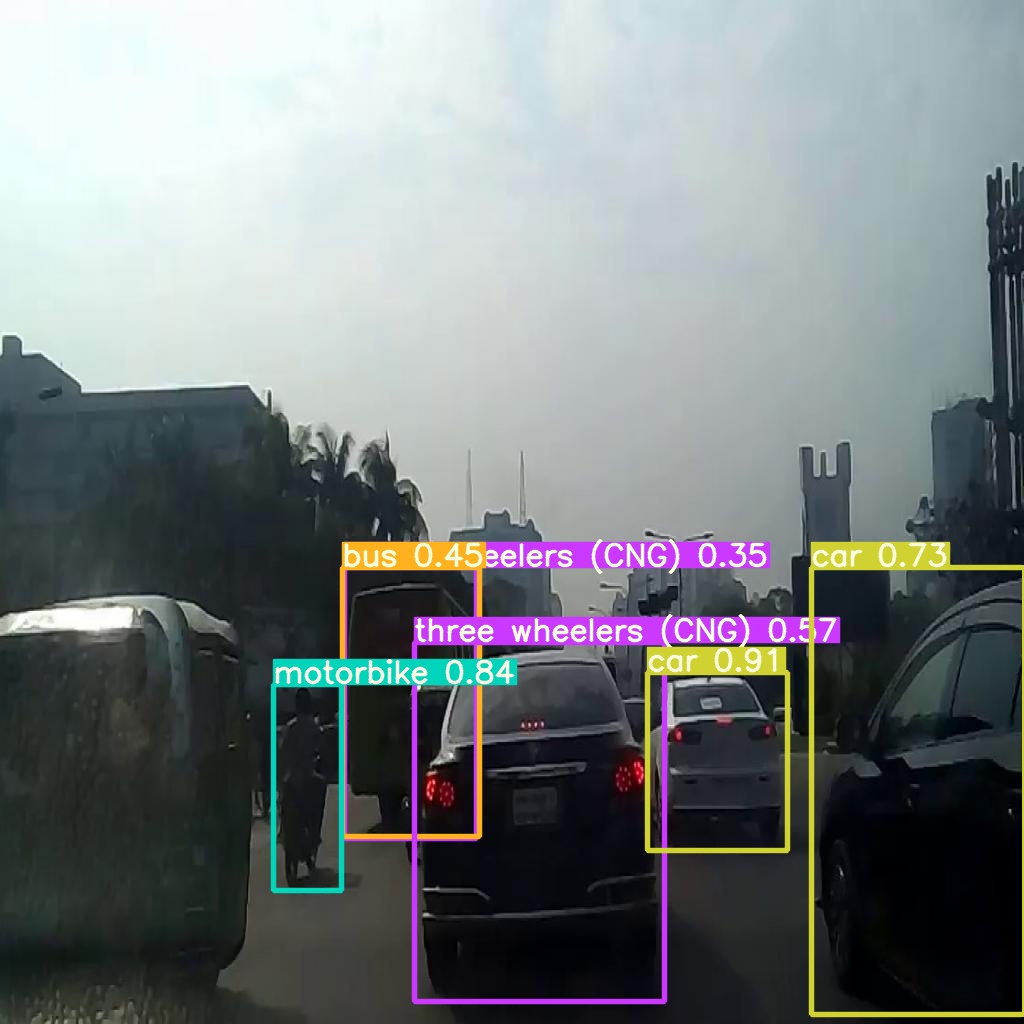}} \\

    (b)YOLOv5s\\

    \hspace{0.5pt}

    % \subfloat[][ \label{...}]
    {\includegraphics[width=0.24\linewidth]{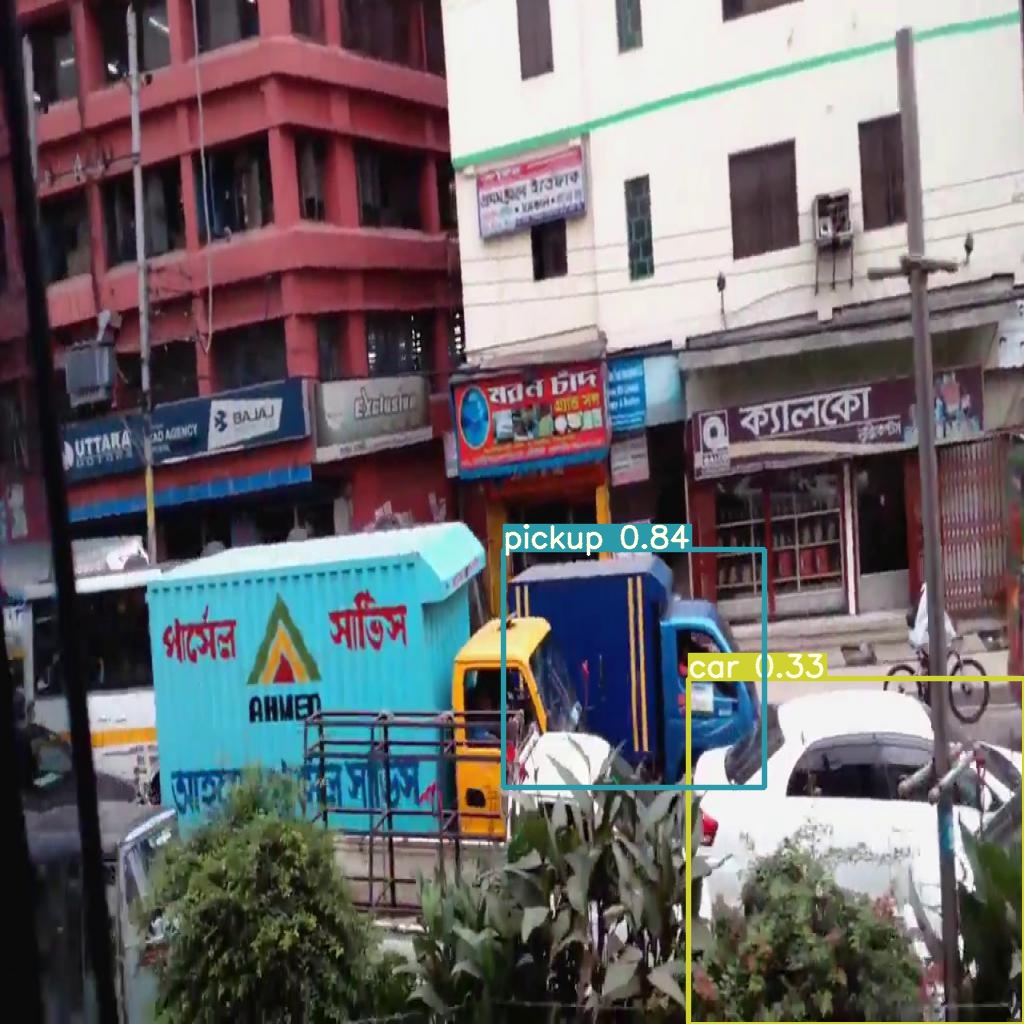}}
    % \subfloat[][ \label{...}]
    {\includegraphics[width=0.24\linewidth]{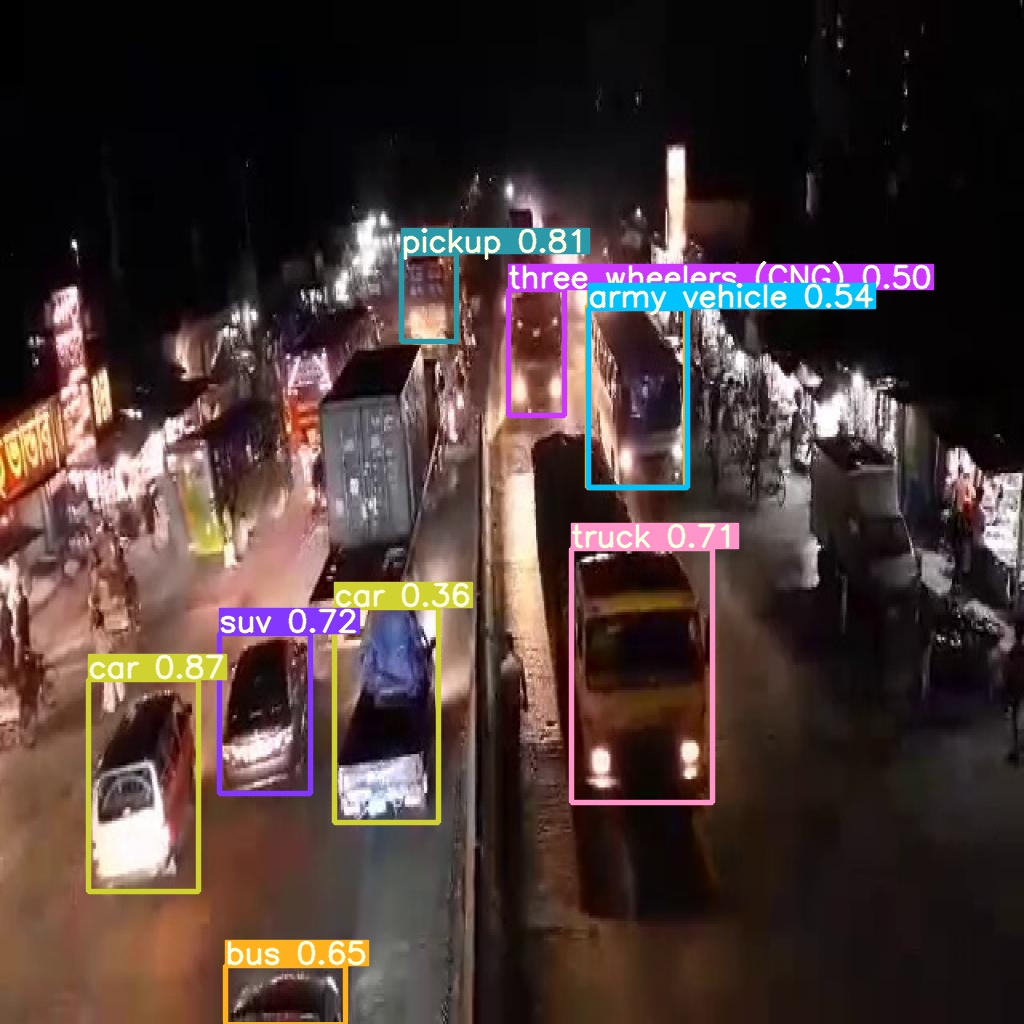}}
    % \subfloat[][ \label{...}]
    {\includegraphics[width=0.24\linewidth]{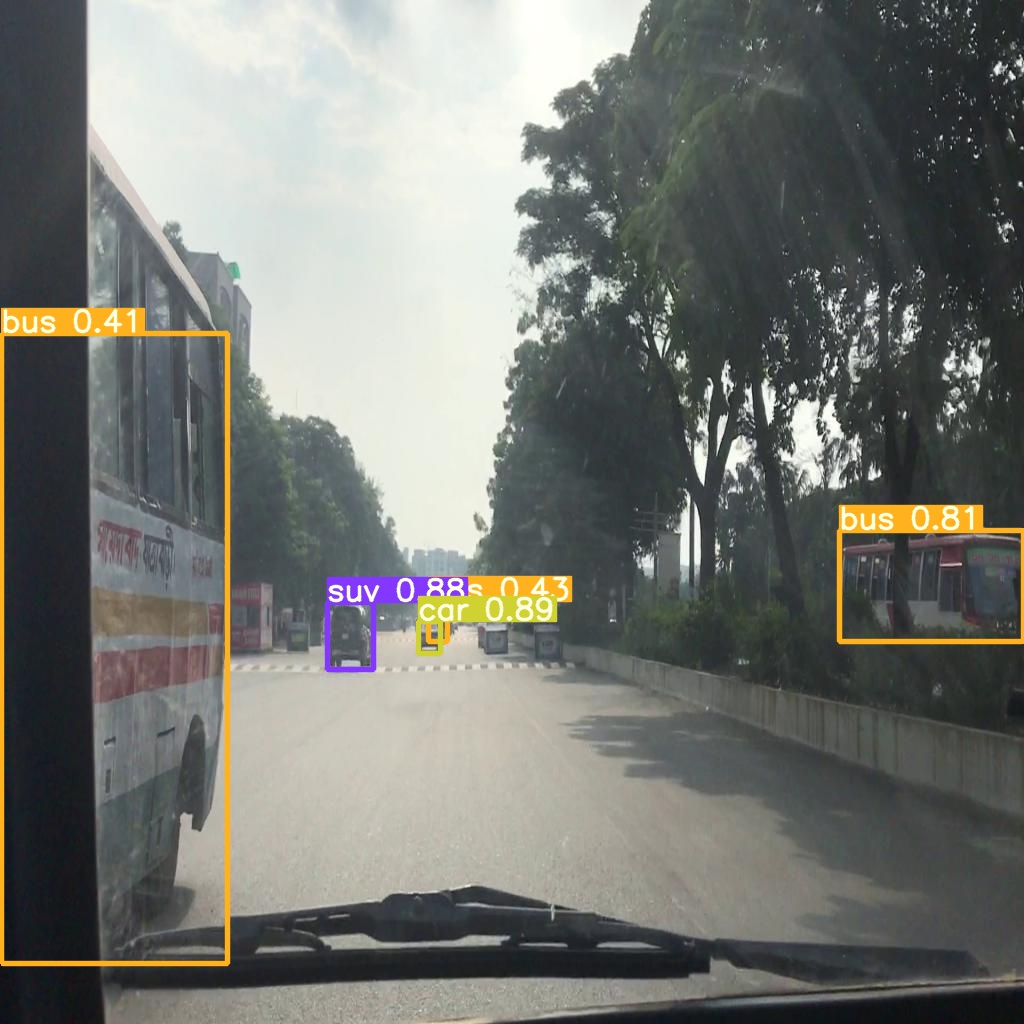}}
    %\subfloat[][YOLOV5X \label{...}]
    {\includegraphics[width=0.24\linewidth]{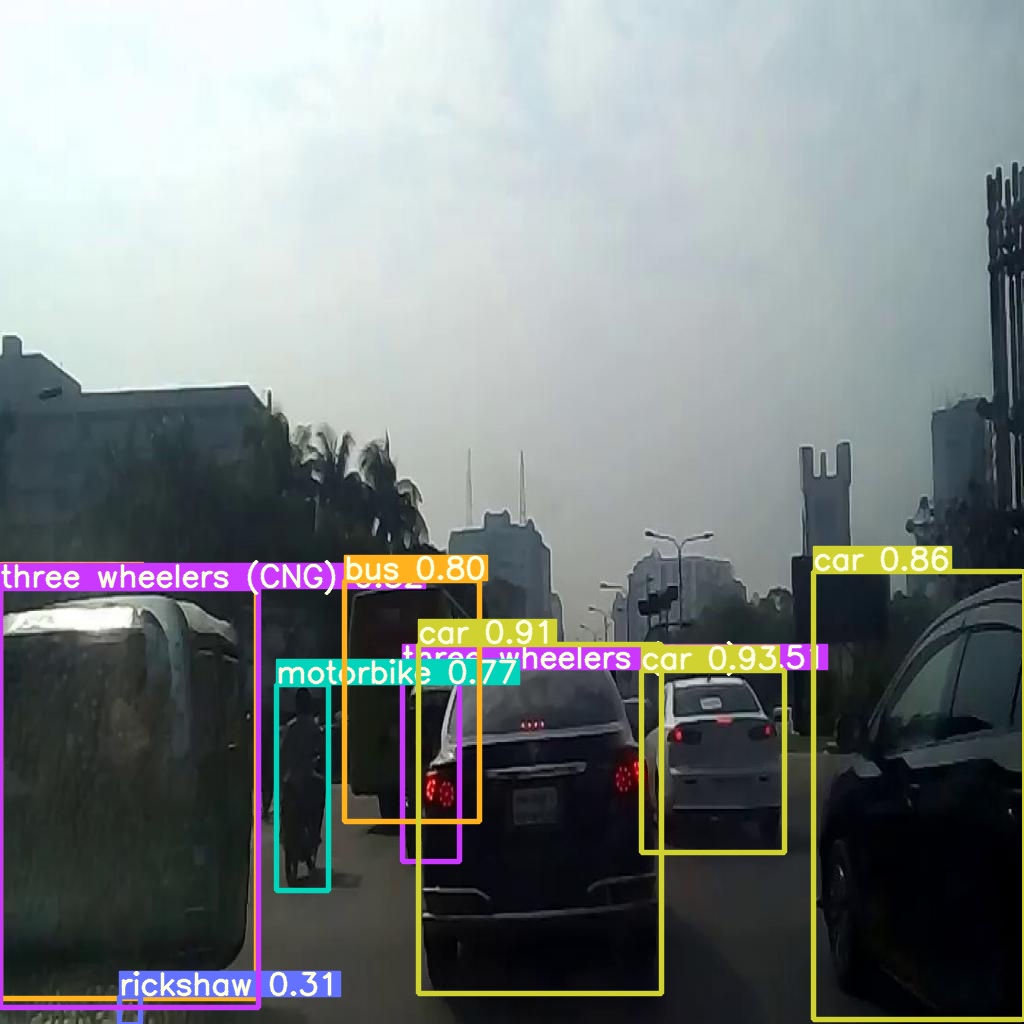}} \\
    
    (c)YOLOv5x\\
    
    \caption{Comparisons between different object detection models}
    \label{fig:comparisons}
 
\end{figure*}

Analyzing Figure~\ref{fig:comparisons} we can see that, in the first example which is a sidewise image of traffic vehicles, it is seen that YOLOv5x can correctly classify most of the vehicles. YOLOv5s was able to detect the vehicles but mistakes were made in its classification. Again, YOLOv3 fails to detect many of the vehicles. 
A similar scenario is seen when night-time images, images of vehicles that are far away, and images from behind are used for inference. YOLOv5x seems to be ahead of the two models with the highest number of correct classifications. The model is followed by YOLOv5s and YOLOv3. YOLOv5s has a slight edge ahead of YOLOv3 as it can detect vehicles better than the latter.

After comparing the different models, it is obvious that YOLOv5x provides better results. It is comparatively accurate and much easier to handle for larger datasets. Bounding boxes with their confidence or probability of the highest probable class are generated. If we take a look at Figure \ref{fig:comparisons}, we can see that YOLOv5x has outperformed all the other models in all the scenarios. The images taken from the sides are accurately used by the model for detection and classification. Images taken from above at night are much for difficult for detection algorithms. YOLOv5x has a better performance here as well in comparison to the other models. Partially visible vehicles resemble other vehicles in many cases. For example, the front of the rickshaw is quite similar to that of a bicycle. The front of many different classes of motor vehicles is similar. YOLOv5x performed better than others when vehicles are partially visible in the image. Similar-looking vehicles are also distinguished more accurately by YOLOv5x.

\subsection{Error Analysis}

\begin{figure}[htbp]
    \centering
    \subfloat[][Minivan undetected \label{YOLOv5x_bad_minivanundetected}]
    {\includegraphics[width=0.42\linewidth]{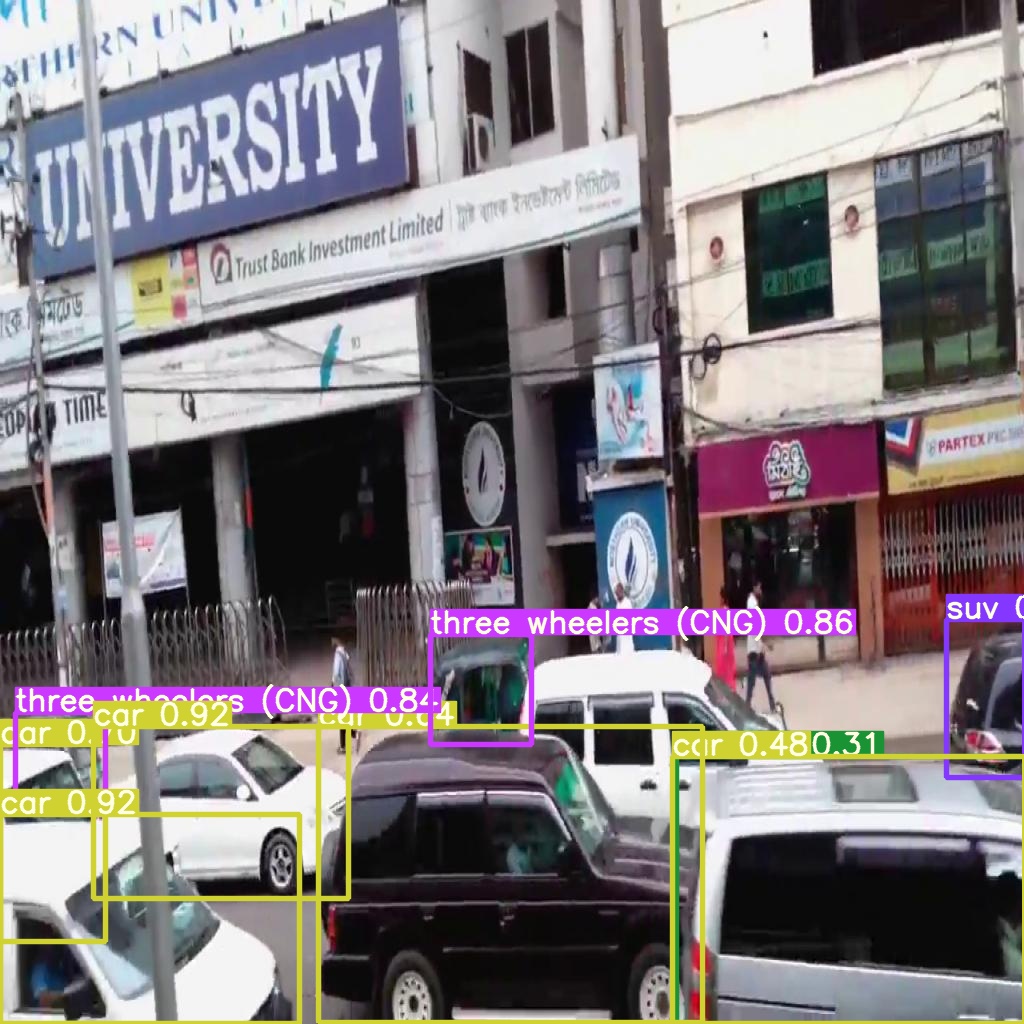}} \quad
    \subfloat[][Vehicles undetected at nighttime \label{YOLOv5x_bad_nighttime}]
    {\includegraphics[width=0.42\linewidth]{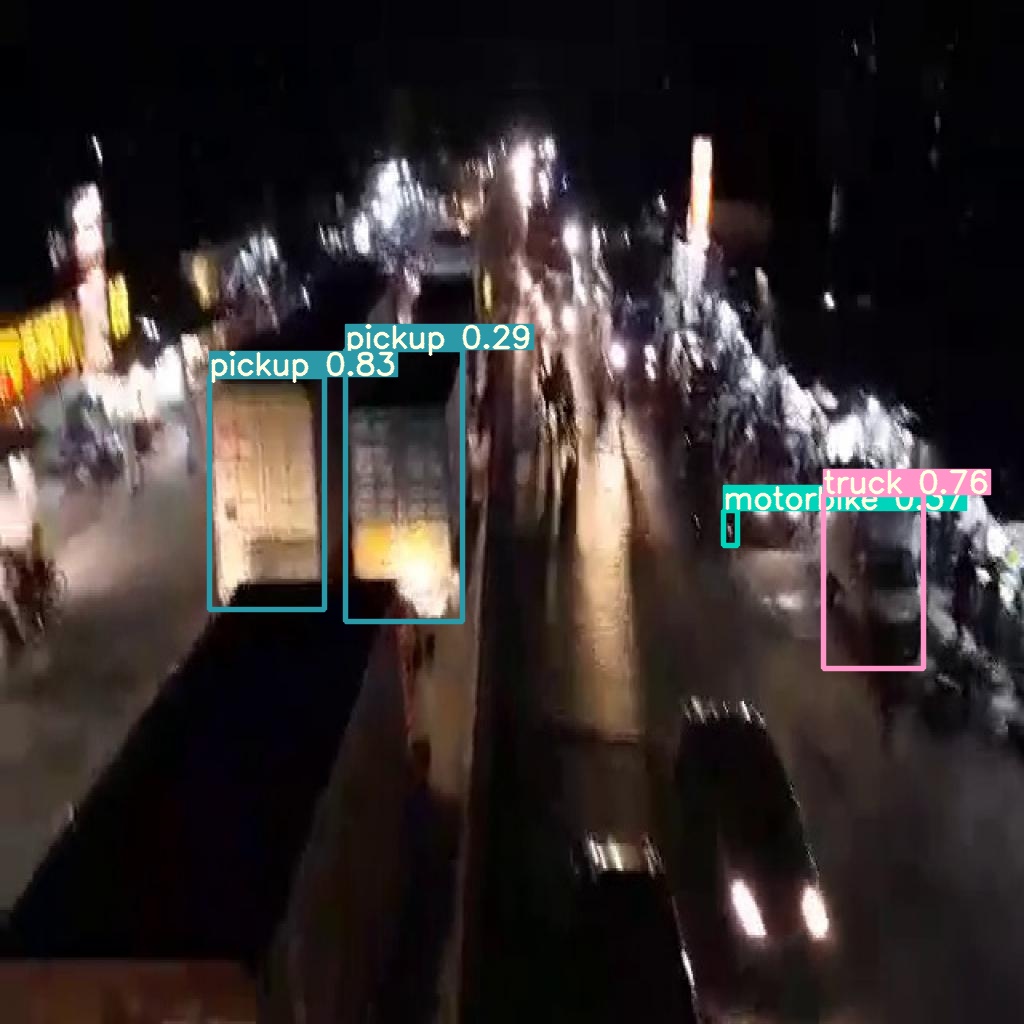}}\\
    \subfloat[][Partially visible vehicles undetected \label{YOLOv5x_bad_partiallyvisiblecar}]
    {\includegraphics[width=0.42\linewidth]{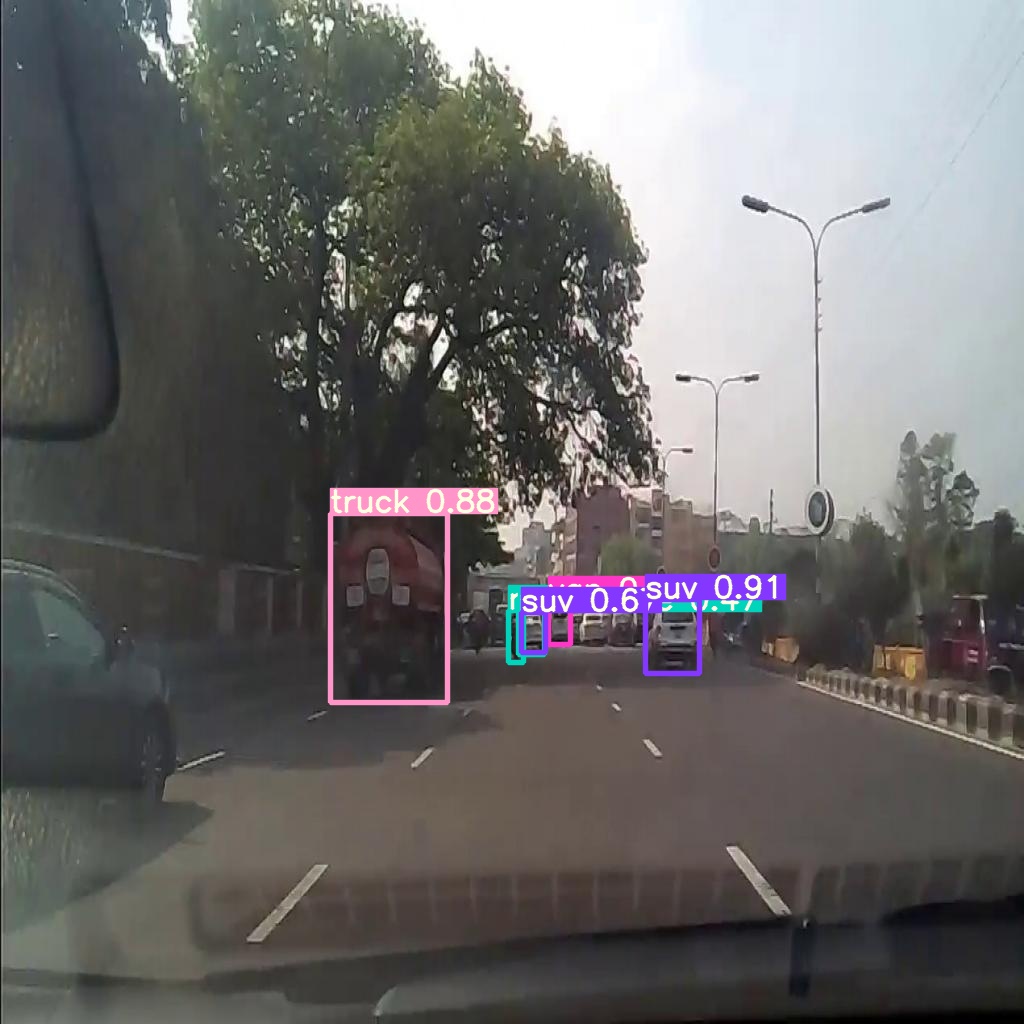}} \quad
    \subfloat[][Vehicles viewed at an angle misclassified \label{YOLOv5x_bad_viewedatanangle}]
    {\includegraphics[width=0.42\linewidth]{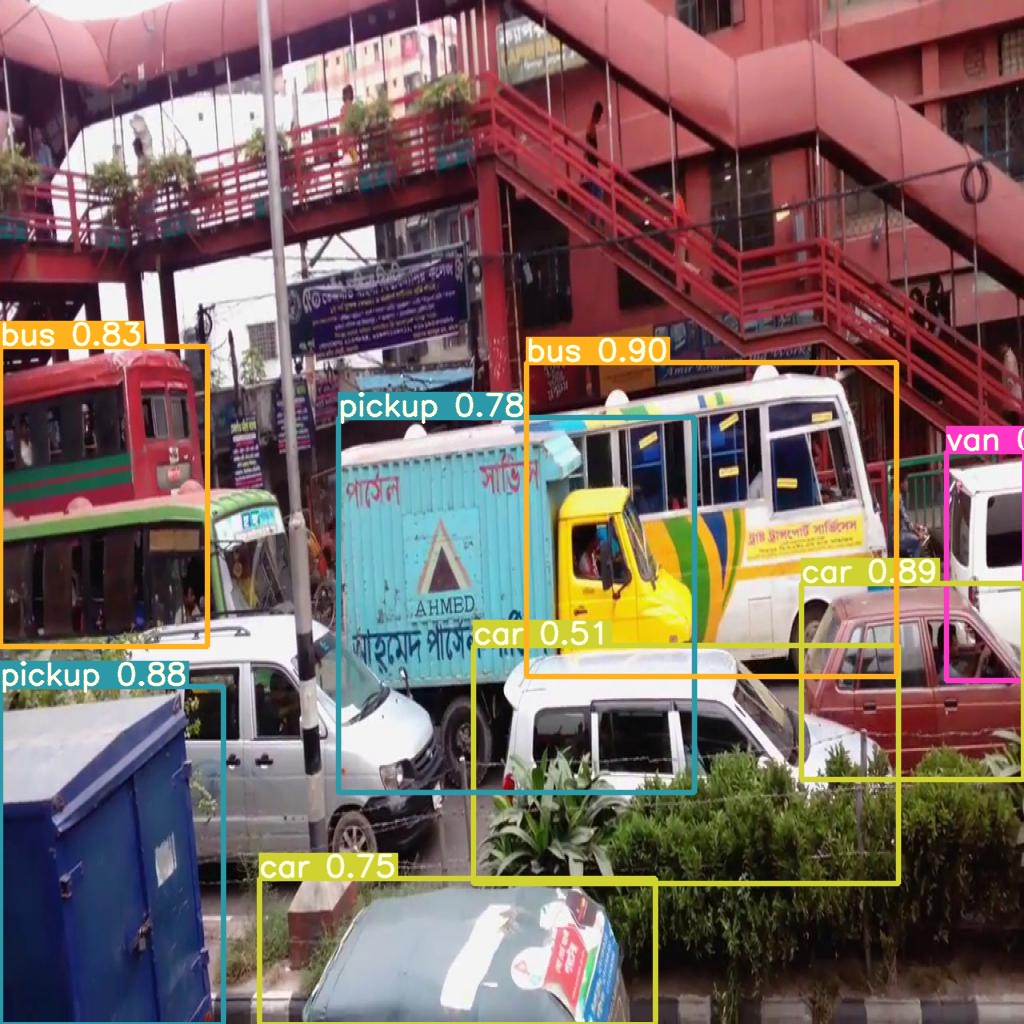}}\\
    
    \caption{Some of the discrepancies observed in YOLOv5x}
    \label{fig:YOLOv5x_bad}
    \vspace*{-0.15cm}
 
\end{figure}

Even though YOLOv5x seems to give impressive results, there are a few flaws noticed during inference. These discrepancies, though rare, can be seen in Figure \ref{fig:YOLOv5x_bad}.
Some specific classes are less detected than others. The reasons are the classes that showed lower performances have been observed to have fewer data to train with and the vehicles also seem to be quite similar to some other classes which resulted in misclassification. For example, a minivan is very similar to an SUV. The performances of the models decrease at night time or during the lack of proper lighting. This can be explained because are visually less noticeable during these conditions. Furthermore, most of the images which were taken in the dark are from a unique position (a foot over-bridge). Vehicles that are partially visible in the frame sometimes get undetected or misclassified. This completely depends on the portion of the vehicle which is in the frame. If there is no significant or detectable part visible, it remains undetected. Sometimes the partial vehicle resembles other types of vehicles. For example, the front of the rickshaw is quite similar to that of a bicycle. Vehicles viewed at an angle are seen to be misclassified or detected with lesser confidence. Fewer training data is the main reason for this discrepancy. Image augmentation has fixed this problem in many cases.

\section{Conclusion and Future Works} \label{sec:conclusion}
In this work, we have evaluated mainly three versions of YOLO-based object detection models for detecting Bangladeshi vehicles. 
The performance of YOLOv3, YOLOv5s, and YOLOv5x have been compared under several working conditions, and YOLOv5x has been the best performer dominating in both overall and class-wise scores.
% We dove deep into their architectures figuring out strengths, weaknesses, and necessary modifications at the implementation level. Afterward, we trained the selected models and presented comparative analyses among them both in terms of quantitative and qualitative.
We have also collected a large traffic dataset consisting of Bangladeshi vehicles from the existing public datasets and our self-exploration. The dataset can be further enriched by including images of different angles, of the night or low light, and of the classes with smaller samples. 
%Our findings, in short, show YOLOv5x to be the best performer. Among the common mistakes and underlying challenges found include - minute differences among various classes, vehicles appearing at a distance or an angle, poor performance in low light, etc.   
In the future, new variants of YOLO-based models, such as YOLOv6 and YOLOv7, may emerge having even better performance than the models we studied here. We incrementally plan to include these models including the ones from different types of architectures, along with more samples and classes in the dataset and more analytical metrics for even scrupulous evaluation.

\bibliographystyle{IEEEtran}
\bibliography{bibfile}

\begin{thebibliography}{10}
\providecommand{\url}[1]{#1}
\csname url@samestyle\endcsname
\providecommand{\newblock}{\relax}
\providecommand{\bibinfo}[2]{#2}
\providecommand{\BIBentrySTDinterwordspacing}{\spaceskip=0pt\relax}
\providecommand{\BIBentryALTinterwordstretchfactor}{4}
\providecommand{\BIBentryALTinterwordspacing}{\spaceskip=\fontdimen2\font plus
\BIBentryALTinterwordstretchfactor\fontdimen3\font minus
  \fontdimen4\font\relax}
\providecommand{\BIBforeignlanguage}[2]{{%
\expandafter\ifx\csname l@#1\endcsname\relax
\typeout{** WARNING: IEEEtran.bst: No hyphenation pattern has been}%
\typeout{** loaded for the language `#1'. Using the pattern for}%
\typeout{** the default language instead.}%
\else
\language=\csname l@#1\endcsname
\fi
#2}}
\providecommand{\BIBdecl}{\relax}
\BIBdecl

\bibitem{hadi2014vehicle}
R.~A. Hadi, G.~Sulong, and L.~E. George, ``Vehicle detection and tracking
  techniques: a concise review,'' \emph{arXiv preprint arXiv:1410.5894}, 2014.

\bibitem{mandal2020object}
V.~Mandal and Y.~Adu-Gyamfi, ``Object detection and tracking algorithms for
  vehicle counting: A comparative analysis,'' \emph{Journal of Big Data
  Analytics in Transportation}, pp. 1--11, 2020.

\bibitem{chintalacheruvu2012video}
N.~Chintalacheruvu, V.~Muthukumar \emph{et~al.}, ``Video based vehicle
  detection and its application in intelligent transportation systems,''
  \emph{Journal of transportation technologies}, vol.~2, no.~04, p. 305, 2012.

\bibitem{Ashrafee_2022_WACV}
A.~Ashrafee, A.~M. Khan, M.~S. Irbaz, and M.~A. Al~Nasim, ``Real-time bangla
  license plate recognition system for low resource video-based applications,''
  in \emph{Proceedings of the IEEE/CVF Winter Conference on Applications of
  Computer Vision (WACV) Workshops}, January 2022, pp. 479--488.

\bibitem{ashikur2022twoDecades}
A.~B.~M. Ashikur~Rahman, M.~B. Hasan, S.~Ahmed, T.~Ahmed, M.~H. Ashmafee, M.~R.
  Kabir, and M.~H. Kabir, ``Two decades of bengali handwritten digit
  recognition: A survey,'' \emph{IEEE Access}, vol.~10, pp. 92\,597--92\,632,
  2022.

\bibitem{fan2016closer}
Q.~Fan, L.~Brown, and J.~Smith, ``A closer look at faster r-cnn for vehicle
  detection,'' in \emph{2016 IEEE intelligent vehicles symposium (IV)}.\hskip
  1em plus 0.5em minus 0.4em\relax IEEE, 2016, pp. 124--129.

\bibitem{zhou2016image}
Y.~Zhou, H.~Nejati, T.-T. Do, N.-M. Cheung, and L.~Cheah, ``Image-based vehicle
  analysis using deep neural network: A systematic study,'' in \emph{2016 IEEE
  international conference on digital signal processing (DSP)}.\hskip 1em plus
  0.5em minus 0.4em\relax IEEE, 2016, pp. 276--280.

\bibitem{zhang2019vehicle}
F.~Zhang, C.~Li, and F.~Yang, ``Vehicle detection in urban traffic surveillance
  images based on convolutional neural networks with feature concatenation,''
  \emph{Sensors}, vol.~19, no.~3, p. 594, 2019.

\bibitem{kasper2021detecting}
M.~Kasper-Eulaers, N.~Hahn, S.~Berger, T.~Sebulonsen, {\O}.~Myrland, and P.~E.
  Kummervold, ``Detecting heavy goods vehicles in rest areas in winter
  conditions using yolov5,'' \emph{Algorithms}, vol.~14, no.~4, p. 114, 2021.

\bibitem{rahman2021densely}
R.~Rahman, Z.~Bin~Azad, and M.~Bakhtiar~Hasan, ``Densely-populated traffic
  detection using yolov5 and non-maximum suppression ensembling,'' in
  \emph{Proceedings of the International Conference on Big Data, IoT, and
  Machine Learning}.\hskip 1em plus 0.5em minus 0.4em\relax Springer Singapore,
  2022, pp. 567--578.

\bibitem{fessel_2019}
K.~Fessel, ``5 {S}ignificant {O}bject {D}etection {C}hallenges and {S}olutions
  --- towardsdatascience.com,''
  \url{https://towardsdatascience.com/5-significant-object-detection-challenges-and-solutions-924cb09de9dd},
  [Accessed 09-Nov-2022].

\bibitem{DhakaAI2020}
\BIBentryALTinterwordspacing
A.~Shihavuddin and M.~R.~A. Rashid, ``{DhakaAI},'' 2020. [Online]. Available:
  \url{https://doi.org/10.7910/DVN/POREXF}
\BIBentrySTDinterwordspacing

\bibitem{tabassum2020poribohon}
S.~Tabassum, S.~Ullah, N.~H. Al-Nur, and S.~Shatabda, ``Poribohon-bd:
  Bangladeshi local vehicle image dataset with annotation for classification,''
  \emph{Data in brief}, vol.~33, p. 106465, 2020.

\bibitem{lowe1999object}
D.~G. Lowe, ``Object recognition from local scale-invariant features,'' in
  \emph{Proceedings of the seventh IEEE international conference on computer
  vision}, vol.~2.\hskip 1em plus 0.5em minus 0.4em\relax Ieee, 1999, pp.
  1150--1157.

\bibitem{viola2001rapid}
P.~Viola and M.~Jones, ``Rapid object detection using a boosted cascade of
  simple features,'' in \emph{Proceedings of the 2001 IEEE computer society
  conference on computer vision and pattern recognition. CVPR 2001},
  vol.~1.\hskip 1em plus 0.5em minus 0.4em\relax IEEE, 2001, pp. I--I.

\bibitem{dalal2005histograms}
N.~Dalal and B.~Triggs, ``Histograms of oriented gradients for human
  detection,'' in \emph{2005 IEEE Computer Society Conference on Computer
  Vision and Pattern Recognition (CVPR'05)}, vol.~1, 2005, pp. 886--893 vol. 1.

\bibitem{felzenszwalb2009object}
P.~F. Felzenszwalb, R.~B. Girshick, D.~McAllester, and D.~Ramanan, ``Object
  detection with discriminatively trained part-based models,'' \emph{IEEE
  transactions on pattern analysis and machine intelligence}, vol.~32, no.~9,
  pp. 1627--1645, 2009.

\bibitem{wei2019multi}
Y.~Wei, Q.~Tian, J.~Guo, W.~Huang, and J.~Cao, ``Multi-vehicle detection
  algorithm through combining harr and hog features,'' \emph{Mathematics and
  Computers in Simulation}, vol. 155, pp. 130--145, 2019.

\bibitem{neumann2017online}
D.~Neumann, T.~Langner, F.~Ulbrich, D.~Spitta, and D.~Goehring, ``Online
  vehicle detection using haar-like, lbp and hog feature based image
  classifiers with stereo vision preselection,'' in \emph{2017 IEEE Intelligent
  Vehicles Symposium (IV)}.\hskip 1em plus 0.5em minus 0.4em\relax IEEE, 2017,
  pp. 773--778.

\bibitem{ahmed2022lessIsMore}
S.~Ahmed, M.~B. Hasan, T.~Ahmed, M.~R.~K. Sony, and M.~H. Kabir, ``Less is
  more: Lighter and faster deep neural architecture for tomato leaf disease
  classification,'' \emph{IEEE Access}, vol.~10, pp. 68\,868--68\,884, 2022.

\bibitem{ren2016faster}
S.~Ren, K.~He, R.~Girshick, and J.~Sun, ``Faster r-cnn: towards real-time
  object detection with region proposal networks,'' \emph{IEEE transactions on
  pattern analysis and machine intelligence}, vol.~39, no.~6, pp. 1137--1149,
  2016.

\bibitem{thuan2021evolution}
D.~Thuan, ``Evolution of yolo algorithm and yolov5: The state-of-the-art object
  detection algorithm,'' \emph{Unpublished undergraduate thesis}, 2021.

\bibitem{Redmon_2016_CVPR}
J.~Redmon, S.~Divvala, R.~Girshick, and A.~Farhadi, ``You only look once:
  Unified, real-time object detection,'' in \emph{Proceedings of the IEEE
  Conference on Computer Vision and Pattern Recognition (CVPR)}, June 2016.

\bibitem{Diwan2022}
T.~Diwan, G.~Anirudh, and J.~V. Tembhurne, ``Object detection using yolo:
  challenges, architectural successors, datasets and applications,''
  \emph{Multimedia Tools and Applications}, Aug 2022.

\bibitem{liu2021performance}
K.~Liu, H.~Tang, S.~He, Q.~Yu, Y.~Xiong, and N.~Wang, ``Performance validation
  of yolo variants for object detection,'' in \emph{Proceedings of the 2021
  International Conference on Bioinformatics and Intelligent Computing}, 2021,
  pp. 239--243.

\end{thebibliography}

\end{document}